%% file: main.tex
\pdfoutput=1
\documentclass[11pt]{article}

\usepackage[]{acl}

\usepackage{times}
\usepackage{latexsym}

\usepackage[T1]{fontenc}
\usepackage[utf8]{inputenc}

\usepackage{microtype}

\usepackage{subcaption}
\usepackage{graphicx}
\usepackage{booktabs}
\usepackage{multicol}
\usepackage{multirow}
\usepackage{amsmath}
\usepackage{amssymb}
\usepackage{bbding}

\input{settings.tex}

\input{math_commands.tex}

\newcommand{\ours}{\textsc{StableMoE}}

\title{\ours{}: Stable Routing Strategy for Mixture of Experts}

\author{
Damai Dai$^{\dag\ddag}$\thanks{\ \  Contribution during internship at Microsoft Research.},~~Li Dong$^\ddag$,~~Shuming Ma$^\ddag$,~~Bo Zheng$^\ddag$,\\
\textbf{Zhifang Sui}$^\dag$,~~\textbf{Baobao Chang}$^\dag$,~~\textbf{Furu Wei}$^\ddag$\\
$^\dag$MOE Key Lab of Computational Linguistics, Peking University \\
$^\ddag$Microsoft Research \\
\texttt{\{daidamai,szf,chbb\}@pku.edu.cn}
\\
\texttt{\{lidong1,shumma,v-zhebo,fuwei\}@microsoft.com} \\
}

\begin{document}

\maketitle

\begin{abstract}
The Mixture-of-Experts~(MoE) technique can scale up the model size of Transformers with an affordable computational overhead.
We point out that existing learning-to-route MoE methods suffer from the routing fluctuation issue, i.e., the target expert of the same input may change along with training, but only one expert will be activated for the input during inference. 
The routing fluctuation tends to harm sample efficiency because the same input updates different experts but only one is finally used. 
In this paper, we propose \textbf{\ours{}} with two training stages to address the routing fluctuation problem.
In the first training stage, we learn a balanced and cohesive routing strategy and distill it into a lightweight router decoupled from the backbone model. 
In the second training stage, we utilize the distilled router to determine the token-to-expert assignment and freeze it for a stable routing strategy. 
We validate our method on language modeling and multilingual machine translation. 
The results show that \ours{} outperforms existing MoE methods in terms of both convergence speed and performance. 
The code is available at \url{https://github.com/Hunter-DDM/stablemoe}. 
\end{abstract}

\section{Introduction}

% [introduction to MoE]
In recent years, large-scale Transformers~\citep{bert,unilm,t5,electra,unilmv2,gpt3} have shown a striking ability to model languages. 
However, with the model scale growing, the training speed will go slower, and the extremely large memory requirement also introduces a heavy burden of engineering. 
Mixture of Experts (MoE)~\citep{ori_moe1,ori_moe2,moe}, in a much easier way, enables Transformers to scale up the number of parameters meanwhile introducing an affordable computational overhead. 
MoE-based Transformers have a set of expert modules, and only a few experts will be activated for each input token. 
In this way, we can expand the model scale by adding expert modules, which will keep the computational and memory overhead within a tolerable range. 

\begin{figure*}[t]
\centering
\includegraphics[width=0.96\linewidth]{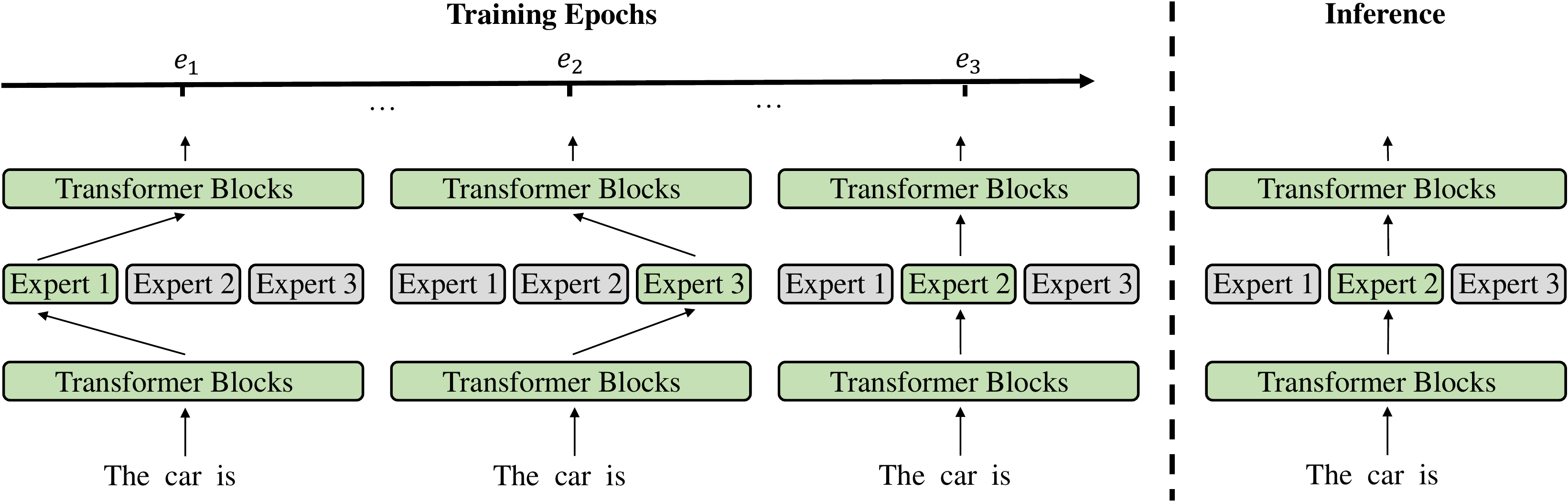}
\caption{
Illustration of the routing fluctuation problem. 
The same input is assigned to different experts along with training. 
However, during inference, only one expert is sparsely activated for the input.
The routing fluctuation tends to harm sample efficiency because the same input updates different experts while only one is used.
}
\label{fig:fluctuation}
\end{figure*}

% [the routing fluctuation problem. + illustration]
Most existing MoE methods~\citep{gshard,switch,base} decide the token-to-expert routing according to the dynamically changing token representations. 
However, we point out that they face the routing fluctuation problem. 
As shown in Figure~\ref{fig:fluctuation}, the same input may be assigned to different experts along with training. 
However, during inference, only one expert will be activated for the input.
The routing fluctuation problem tends to harm sample efficiency because the same input updates different experts while only one is finally used.

\begin{figure}[t]
\centering
\includegraphics[width=0.99\linewidth]{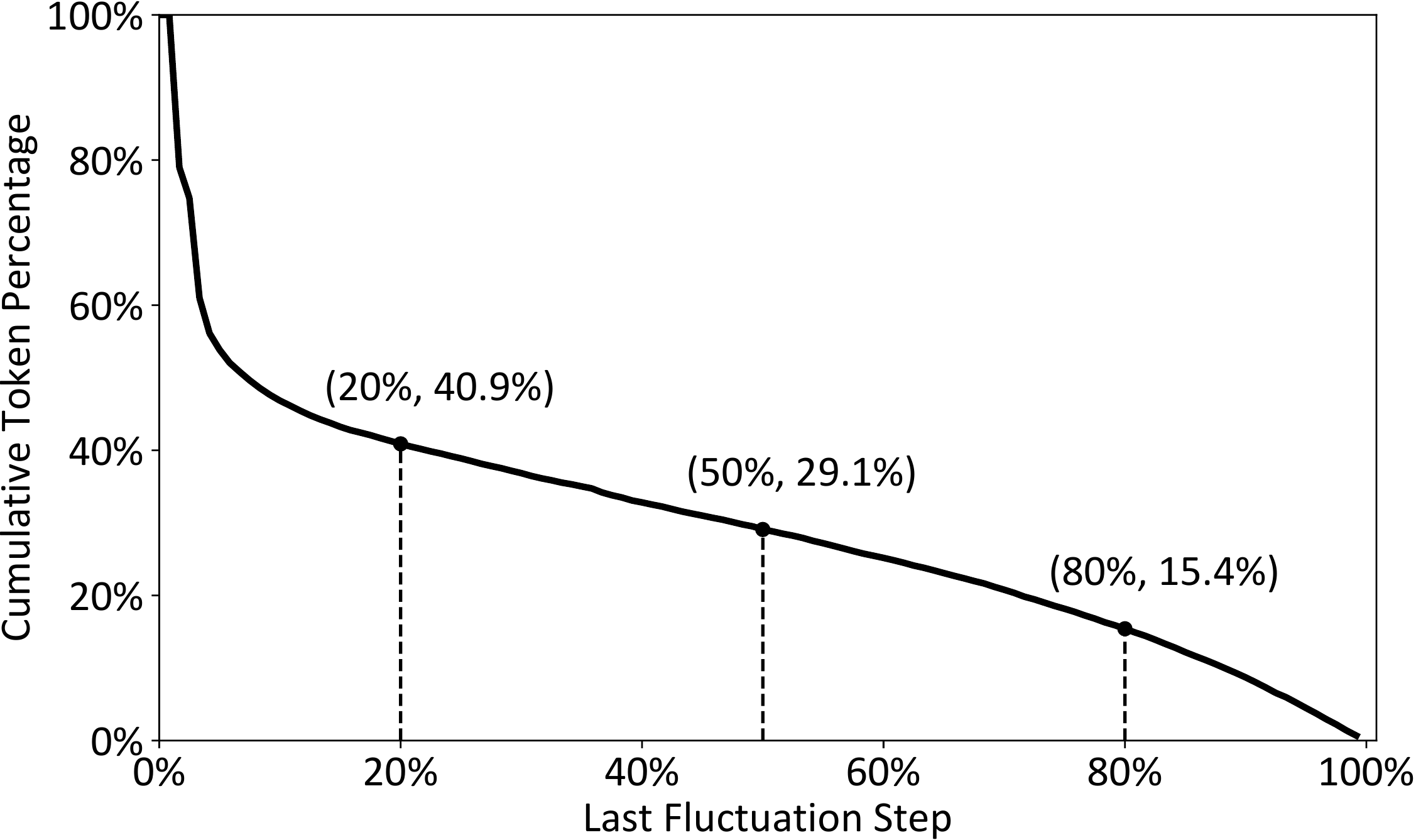}
\caption{
Cumulative token percentage with regard to the last fluctuation step of tokens for BASE Layer~\citep{base}.
%  in the validation set
A substantial portion of tokens still change their target experts even if the training is nearing the end. 
}
\label{fig:last_fluctuation}
\end{figure}

Taking BASE Layer~\citep{base} as an example, during the whole training process, we examine the token-to-expert assignment for tokens in the validation set. 
For an input token, we define the last fluctuation step as the last step where its target expert is different from the final step. 
We plot the cumulative token percentage with regard to the last fluctuation step (annotated as its percentage accounting for all training steps) in Figure~\ref{fig:last_fluctuation}. 
We find that the last fluctuation step of 40.9\% tokens exceeds 20\%, which means 40.9\% tokens do not have a stable target expert when 20\% of all training steps have been done. 
Furthermore, 29.1\% tokens still change their target experts after half of the whole training process, and 15.4\% tokens even change the target expert after 80\% of all training steps, which is nearing the training ending. 
These statistics prove that the routing fluctuation problem indeed exists in previous MoE methods. 

% [we propose \ours{} to address routing fluctuation]
In this paper, we propose \textbf{\ours{}} with two training stages to address the routing fluctuation problem. 
In the first training stage, we follow the learning-to-route paradigm and aim to learn a balanced and cohesive routing strategy. 
We design a balance loss to guarantee the assignment is balanced. 
In addition, inspired by \citet{base}, we adopt a sigmoid gating mechanism, which enables the task objective to propagate supervised signal back to the routing strategy, to facilitate learning a more cohesive assignment. 
As the routing strategy is being learned, we synchronously distill it into a lightweight router decoupled from the backbone model. 
In the second training stage, we utilize the distilled router to determine the token-to-expert assignment. 
The distilled router is frozen in this stage to provide a stable routing strategy, which addresses the routing fluctuation problem in the remaining training. 
% [summary to experimental results]
We conduct experiments on language modeling and multilingual machine translation.
The results show that \ours{} outperforms existing MoE methods in terms of both convergence speed and performance.

Our contributions are summarized as follows: 
(1) We point out the routing fluctuation problem in existing learning-to-route MoE methods. 
(2) We propose \ours{} to address the routing fluctuation problem. 
(3) We conduct substantial experiments under various settings to show the advantages of~\ours{} over existing MoE methods. 

\begin{figure*}[t]
\centering
\includegraphics[width=0.96\linewidth]{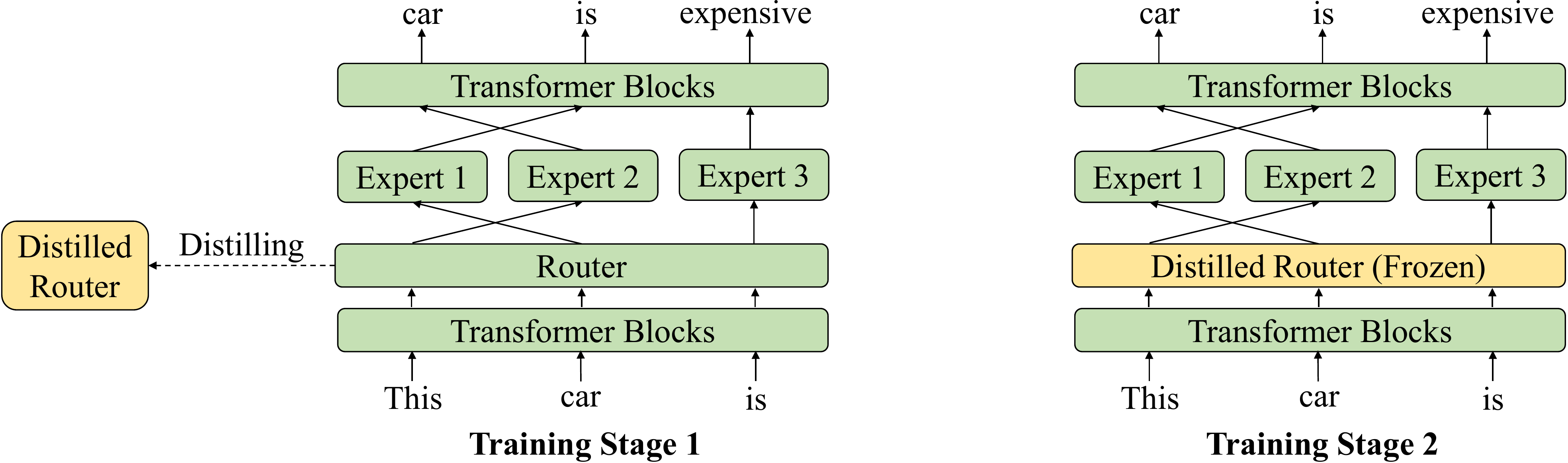}
\caption{
Illustration of two training stages in \ours{}. 
In training stage 1, we learn a routing strategy and distill it into a lightweight router. 
Then, we freeze the distilled router for stable routing in training stage 2. 
}
\label{fig:stable_moe}
\end{figure*}

\section{Background: Mixture-of-Experts for Transformers}

% [introduction to MoE Transformers]
We first introduce the MoE mechanism designed for Transformers~\citep{transformer}. 
Given a standard $L$-layer Transformer model and an input sequence $X$ containing $T$ tokens, the Transformer output $H^{L}$ is calculated by
\begin{align}
    H^{L} &= [\mathbf{h}^{L}_1; \mathbf{h}^{L}_2; ...; \mathbf{h}^{L}_T], \\
    \mathbf{h}_{t}^{l} &= \operatorname{FFN}\left( \mathbf{u}_{t}^{l} \right) + \mathbf{u}_{t}^{l}, \\
    \mathbf{u}_{1:T}^{l} &= \operatorname{self-att}\left( \mathbf{h}_{1:T}^{l-1} \right) + \mathbf{h}_{1:T}^{l-1},
\end{align}
where $\mathbf{h}_{t}^{l}$ is the hidden state of $t$-th token after the $l$-th layer, $\operatorname{Self-Att}(\cdot)$ is the self-attention module, and $\operatorname{FFN}(\cdot)$ is short for the feed-forward network. 
For simplicity, we omit the layer normalization. 

% Following~\citep{base}, w
We implement MoE for Transformers by inserting MoE layers, which are composed of a set of FFNs, into two neighboring Transformer blocks. 
At an MoE layer, for each input token, only a few or one expert will be activated, controlled by a gating function $g(\cdot)$: 
\begin{equation}
\mathbf{h}_{t}^{l} = \sum_{i=1}^{N}{g_{i}\left( \mathbf{h}_{t}^{l-1} \right)\operatorname{FFN}_{i}\left( \mathbf{h}_{t}^{l-1} \right)} + \mathbf{h}_{t}^{l-1},
\end{equation}
where $N$ is the total number of experts, and $\operatorname{FFN}_{i}$ is the $i$-th expert. Here, the gating function $g_{i}(\cdot)$ is sparse for computational efficiency. 
For simplicity, we omit the layer normalization.

\section{Method}

% [introduction to our two-stage framework]
\ours{} has two training stages as illustrated in Figure~\ref{fig:stable_moe}. 
In the first training stage, we follow the learning-to-route paradigm and aim to learn a balanced and cohesive routing strategy. 
As the routing strategy is being learned, we synchronously distill it into a lightweight router decoupled from the backbone model. 
In the second training stage, we utilize the distilled router to determine the token-to-expert assignment. 
The distilled router is frozen in this stage to provide a stable routing strategy. 
During inference, we also use the frozen distilled router for consistent routing. 

\subsection{Training Stage 1: Learn Routing Strategy}

% [assignment and forward]
% At the MoE layer, l
Let $\mathbf{h}_{t}^{l-1} \in \mathbb{R}^{d}$ be the input representation of token $t$ and $E \in \mathbb{R}^{N \times d}$ be the centroids of $N$ experts.
For each MoE layer, we assign each token to one expert FFN~\citep{switch,base,hash}.
The assignment score is:
\begin{equation}
    s_{t, i} = E_i^{\top} \mathbf{h}_{t}^{l-1},
\end{equation}
where $s_{t,i}$ is the assignment score between token $t$ and expert $i$, indicating their affinity. 
We use a greedy assignment algorithm, i.e., sending each token to the expert with the highest affinity. 
Then, we calculate the expert FFN output as:
\begin{align}
    a_t &= \mathop{\arg\max}_{i}(s_{t,i}), \label{equ:at} \\
    \mathbf{h}_{t}^{l} &= \sigmoid\left( s_{t,a_t} \right)\operatorname{FFN}_{a_t}\left( \mathbf{h}_{t}^{l-1} \right) + \mathbf{h}_{t}^{l-1},
    \label{equ:gate_moe}
\end{align}
where $a_t$ is the expert index that token $t$ is sent to, and $\sigmoid$ is the sigmoid gate~\citep{base}.
Considering the sigmoid gate $\sigmoid\left( s_{t,a_t} \right)$, if $\operatorname{FFN}_{a_t}$ is beneficial for token $t$, optimizing the training objective (e.g., minimizing the cross-entropy loss for language modeling) will urge the gate to be greater; otherwise, the gate will tend to be smaller. 
The gate signal urges similar tokens to be assigned to the same expert that is beneficial to them, thus producing cohesive token-to-expert assignments. 

\paragraph{Balance Loss}
We design a balance loss $\mathcal{L}_{bal}$ to avoid imbalanced assignments that will result in a high computational bottleneck in the MoE layer and thus limit the computational efficiency: 
\begin{equation}
    \mathcal{L}_{bal} = \alpha \sum_{i=1}^{N}{\left( \frac{\left(|\mathcal{A}_i| - \overline{n} \right)}{\overline{n}} \sum_{t \in \mathcal{A}_i}{\sigmoid\left( s_{t,i} \right)}\right)}, 
    \label{equ:balance_loss}
\end{equation}
where $\alpha$ is a hyper-parameter, $\mathcal{A}_i$ denotes the set of tokens assigned to expert $i$, and $\overline{n}$ denotes the average number of tokens per expert. 
Intuitively, if an expert is overloaded, the balance loss will urge its assignment scores to be smaller. 
Otherwise, if an expert is unoccupied, the balance loss will increase its assignment scores to capture more tokens. 

\begin{table*}[t]
\centering
% \footnotesize
\small
\setlength{\tabcolsep}{27pt}
\begin{tabular}{@{}l | c c c@{}}
\toprule
\textbf{Methods} & \textbf{Assignment Algorithm} & \textbf{Gating Function} & \textbf{Balance Loss}\\
\midrule
Switch Transformer & Greedy & $\mathrm{softmax}$ & Yes \\
BASE Layer & Auction~\citep{auction} & $\mathrm{sigmoid}$ & No \\
Hash Layer & Fixed Hashing & $\{0, 1\}$ & No \\
\midrule
\ours{} &  &  &  \\
~~~~Training Stage 1 & Greedy & $\mathrm{sigmoid}$ & Yes \\
~~~~Training Stage 2 & Fixed Routing & $\mathrm{sigmoid}$ & No \\
\bottomrule
\end{tabular}
\caption{
Comparison of three core elements among \ours{} and existing MoE-based Transformers. 
}
\label{tab:compare}
\end{table*}

% [distill]
\paragraph{Distilled Router}
As the routing strategy is being learned, we synchronously distill it into a lightweight router decoupled from the backbone model to mimic the original routing strategy. 
Let $X$ be the input sequence and $\hat{E}$ be the distilled expert centroids, we use word embeddings $\operatorname{D}(\cdot)$ to extract the routing features. 
We use the cross-entropy loss as the distillation loss $\mathcal{L}_{dis}$:
\begin{align}
\hat{\mathbf{h}}_{t}^{l-1} &= \operatorname{D}(X_{t}), \quad \hat{s}_{t, i} = \hat{E}_{i}^{\top} \hat{\mathbf{h}}_{t}^{l-1}, \\
\mathcal{L}_{dis} &= -\sum_{t=1}^{T}{\log \frac{\exp \left(\hat{s}_{t, a_t}\right)}{\sum_{i=1}^{N} \exp \left(\hat{s}_{t, i}\right)}},
\end{align}
where $\hat{\mathbf{h}}_{t}^{l-1}$ is the distilled routing feature of token $t$, $\hat{s}_{t, i}$ is the distilled assignment score between token $t$ and expert $i$, and $a_t$ is the expert index that token $t$ is actually sent to. 
In practice, $\operatorname{D}(\cdot)$ can also be other feature extractors such as CNNs or Transformers (we investigate other variants of distilled routers in Section~\ref{sec:distill}), but the word embedding is the fastest one and achieves the best performance. 
At the end of training stage 1, we freeze all parameters for the distilled router (i.e., $\operatorname{D}(\cdot)$ and $\hat{E}$) to prepare a stable routing strategy for training stage 2 and the inference stage. 

\paragraph{Training Objective}
In training stage 1, the training loss consists of the task loss, the balance loss, and the distillation loss: 
\begin{equation}
    \mathcal{L}_{S1} = \mathcal{L}_{task} + \mathcal{L}_{bal} + \mathcal{L}_{dis}. 
\end{equation}

\subsection{Training Stage 2: Learn with Stable Routing Strategy}

% [introduction to training stage 2]
Given frozen $\operatorname{D}(\cdot)$ and $\hat{E}$, in training stage 2, we directly use them for a stable routing strategy. 
Keeping other processes the same as in training stage 1, we calculate the output of the MoE layer as follows: 
\begin{align}
    \hat{\mathbf{h}}_{t}^{l-1} &= \operatorname{D}(X_{t}), \quad \hat{s}_{t, i} = \hat{E}_i^{\top} \hat{\mathbf{h}}_{t}^{l-1}, \\
    \hat{a}_t &= \mathop{\arg\max}_{i}(\hat{s}_{t,i}), \\
    \mathbf{h}_{t}^{l} &= \sigmoid\left( s_{t,\hat{a}_t} \right)\operatorname{FFN}_{\hat{a}_t}\left( \mathbf{h}_{t}^{l-1} \right) + \mathbf{h}_{t}^{l-1}.
\end{align}

Notice that the sigmoid gate $\sigmoid(\cdot)$ still uses original assignment score $s_{t,\hat{a}_t}$ as input, so the gate signal can also be learned in training stage 2. 
Since the routing strategy has been fixed in training stage 2, we no longer need the balance loss and distillation loss. 
Therefore, the training loss for training stage 2 contains only the task loss: 
\begin{equation}
    \mathcal{L}_{S2} = \mathcal{L}_{task}.
\end{equation}

\subsection{Inference}
\label{sec:inference}

During inference, we also use the frozen distilled router for routing. 
The fixed routing strategy, which is consistent with training stage 2, makes information learned in MoE layers be utilized more thoroughly and thus leads to better performance. 

\subsection{Comparison with Existing MoE Methods}

% [comparison with existing methods. + table]
We compare three core elements, including the assignment algorithm, the gating function, and the balance loss, among \ours{} and existing MoE-based Transformers. 
In Table~\ref{tab:compare}, we summarize their differences. 

\paragraph{Assignment Algorithm}
Switch Transformer and the training stage 1 in \ours{} simply assign each token to the expert with the highest affinity. 
BASE Layer adopts the auction algorithm~\citep{auction} to find a global balanced assignment with the maximum affinity sum. 
Hash layer and the training stage 2 in \ours{} have token-level fixed routing strategies, which have good stability. 

\paragraph{Gating Function}
Hash Layer uses a hard gating function, which means an expert is either fully activated or not activated, no any intermediate state. 
Switch Layer, BASE Layer, and \ours{} have soft gating functions, which can judge the affinity between a token and its target expert and determine a proper ratio to use the expert. 
Soft gating mechanisms also urge models to learn a more cohesive token-to-expert assignment. 

\paragraph{Balance Loss}
BASE Layer and Hash Layer do not apply any balance losses. 
By contrast, Switch Transformer and the training stage 1 in \ours{} design balance losses to control the balance of the token-to-expert assignment. 

In summary, combing two training stages, \ours{} has a stable, cohesive, and balanced routing strategy, while the other three MoE methods cannot meet them all simultaneously.

\begin{table*}[t]
\centering
\footnotesize
\setlength{\tabcolsep}{7pt}
\begin{tabular}{@{}c | l | c c c | c c@{}}
\toprule
\textbf{Size} & \textbf{Models} & \textbf{\# Shared Params} & \textbf{\# Expert Params} & \textbf{FLOPs} & \textbf{Valid PPL} & \textbf{Test PPL} \\
\midrule
\multirow{7}{*}{Base} & Standard Transformer & 124M & N/A & 146B & 23.02 & 22.58 \\
& Larger Transformer (deeper) & 578M & N/A & 610B & 17.93 & 17.63 \\
& Larger Transformer (wider) & 578M & N/A & 610B & 18.31 & 18.01 \\
\cmidrule{2-7}
& Switch Transformer & 124M & 454M & 160B & 19.79 & 19.20 \\
& BASE Layer & 124M & 454M & 160B & 20.04 & 19.69 \\
& Hash Layer & 124M & 454M & 160B & 19.63 & 19.25 \\
& \ours{} & 124M & 454M & 160B & \textbf{19.28} & \textbf{18.93} \\
\midrule
\multirow{5}{*}{Large} & Standard Transformer & 355M & N/A & 414B & 18.86 & 18.19 \\
\cmidrule{2-7}
& Switch Transformer & 355M & 3.22B & 465B & 16.62 & 16.21 \\
& BASE Layer & 355M & 3.22B & 465B & 16.36 & 15.75 \\
& Hash Layer & 355M & 3.22B & 465B & 16.37 & 15.79 \\
& \ours{} & 355M & 3.22B & 465B & \textbf{16.22} & \textbf{15.59} \\
\bottomrule
\end{tabular}
\caption{
Perplexity results of language modeling.
We also report the training FLOPs, and the number of parameters for the shared backbone (\# Shared Params) and the expert layers (\# Expert Params).
``N/A'' denotes not applicable.
\ours{} consistently outperforms other MoE methods under both the base and the large settings. 
}
\label{tab:main_lm}
\end{table*}

\section{Experiments}

\subsection{Tasks and Datasets}

% [language modeling, RoBERTa + CC100]
\paragraph{Language Modeling} Following~\citep{base} and~\citet{hash}, we use the combination of the corpora in RoBERTa~\citep{roberta} and the English subset of the CC100~\citep{cc100} corpus. 
The corpus contains about 100B tokens, and we randomly sample 5M tokens for validation and 20M tokens for test. 

% [machine translation, WMT]
\paragraph{Multilingual Machine Translation} We follow~\citet{zcode} and~\citet{xlmt} to use a collection of parallel data in different languages from the WMT datasets.\footnote{http://www.statmt.org} 
The dataset contains 32.5 million parallel data for language pairs between English and other 9 languages, including French (Fr), Czech (Cs), German (De), Finnish (Fi), Latvian (Lv), Estonian (Et), Romanian (Ro), Hindi (Hi), and Turkish (Tr). 
In our experiments, we combine the original parallel data with 180 million back-translation data as described in~\citep{xlmt} and call the augmented dataset WMT for short. 

\subsection{Experimental Setup}

% [General]
We conduct experiments based on fairseq\footnote{https://github.com/facebookresearch/fairseq}.
All experiments are conducted on NVIDIA V100 GPUs with 32 GB memory. 

% [setup of LM]
\paragraph{Language Modeling} 
We adopt the tokenizer of GPT-2~\citep{gpt2}, which uses byte-pair encoding~\citep{bpe} with a vocabulary size of 50,257. 
We set up two settings for \ours{}, a base one and a large one. 
For both settings, we insert one MoE layer after the middle Transformer block. 
We train the model for 60K steps in total (6K for training stage 1 and 54K for training stage 2). 
The dimension of the distilled routing features is 50, which brings 2.51M extra parameters for routing. 
The balance factor $\alpha$ is set to 0.3. 
We use Adam~\citep{adam} with $\beta_1=0.9$ and $\beta_2=0.98$ as the optimizer. 
The rest of the hyper-parameters are summarized in Appendix~\ref{appendix:hyper_lm}. 

% [setup of MT]
\paragraph{Multilingual Machine Translation}
Following~\citep{xlmt}, we use the SentencePiece~\citep{sentence_piece} model to tokenize sentences. 
The vocabulary is learned from the training set and consists of 64,000 tokens. 
We insert two MoE layers, one after the third encoder block and one after the third decoder block. 
We train the model for 352K steps in total (30K for training stage 1 and 322K for training stage 2). 
The dimension of the distilled routing features is also set to 50. 
The balance factor $\alpha$ is set to 0.3. 
We use Adam with $\beta_1=0.9$ and $\beta_2=0.98$ as the optimizer. 
The rest of the hyper-parameters are summarized in Appendix~\ref{appendix:hyper_mt}. 

\subsection{Results}

\begin{table*}[t]
\centering
\footnotesize
\setlength{\tabcolsep}{6pt}
\begin{tabular}{@{}l | c c | c c c c c c c c c | c@{}}
\toprule
\textbf{Models} & \textbf{\# Params} & \textbf{FLOPs} & \textbf{De} & \textbf{Ro} & \textbf{Fr} & \textbf{Cs} & \textbf{Et} & \textbf{Hi} & \textbf{Tr} & \textbf{Fi} & \textbf{Lv} & \textbf{Avg} \\
\midrule
Standard Transformer & 77M & 290B & 39.8 & 36.0 & 32.5 & 29.1 & 27.2 & 24.5 & 23.6 & 21.8 & 20.3 & 28.31 \\
Larger Transformer & 90M & 317B & 40.6 & 36.9 & 33.7 & 29.8 & 27.8 & 25.4 & 24.6 & 22.2 & 20.9 & 29.10 \\
\midrule
Switch Transformer & 480M & 317B & 42.3 & 37.1 & 33.8 & 31.0 & 28.6 & 26.0 & 24.3 & 23.0 & 21.2 & 29.70 \\
BASE Layer & 480M & 317B & 42.6 & \textbf{37.8} & 34.2 & 31.0 & 29.0 & \textbf{26.9} & \textbf{25.1} & 23.2 & 21.6 & 30.16 \\
Hash Layer & 480M & 317B & 42.7 & 37.0 & 34.6 & 31.3 & 28.7 & 26.5 & 23.9 & 23.1 & 21.7 & 29.94 \\
\ours{} & 480M & 317B & \textbf{43.0} & 37.4 & \textbf{34.7} & \textbf{31.5} & \textbf{29.3} & 26.8 & 24.7 & \textbf{23.6} & \textbf{21.9} & \textbf{30.32} \\
\bottomrule
\end{tabular}
\caption{
X$\rightarrow$En test BLEU on WMT.
We also report the total number of parameters, and training FLOPs.
\ours{} outperforms other MoE-based Transformers across most languages. 
}
\label{tab:main_nmt}
\end{table*}

\subsubsection{Language Modeling}

We compare \ours{} with Switch Transformer, BASE Layer, Hash Layer, and the standard Transformer. 
All MoE models have the same number of shared parameters as the standard Transformer. 
Under the base setting, in addition, we compare two larger dense Transformers that add FFNs in a dense manner to achieve the same number of total parameters as MoE models. 
The deeper model stacks more FFNs, while the wider model uses FFNs with a larger hidden size. 
The floating point operations~(FLOPs) per sequence are profiled by the torchprofile toolkit.

We show the main results of language modeling on the RoBERTa+cc100en corpus in Table~\ref{tab:main_lm}. 
Under the base setting, \ours{} outperforms existing MoE methods on both the validation and the test sets by 0.3-0.8 perplexity. 
Compared with dense models, \ours{} achieves about 3.7 lower perplexity than the standard Transformer, and about 1.3 higher perplexity than the deeper larger model. 
Under the large setting, consistently, \ours{} outperforms the other MoE methods, and achieves about 2.6 lower perplexity than the standard Transformer. 

\begin{figure}[t]
\centering
\includegraphics[width=0.96\linewidth]{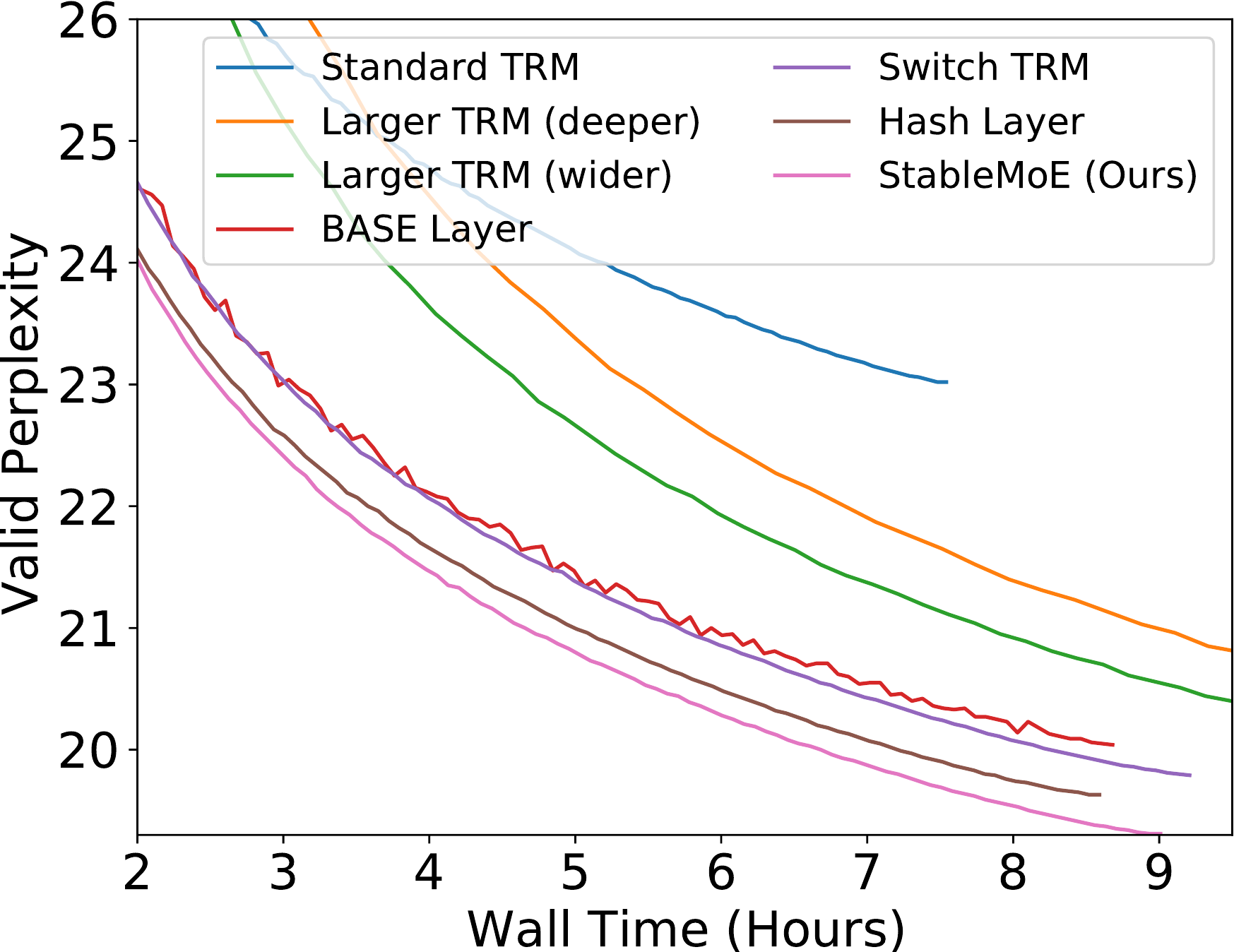}
\caption{
Convergence speed of different models. 
TRM is a shorthand for Transformer. 
}
\label{fig:convergence}
\end{figure}

We also compare the convergence speed of different models under the base setting. 
The results are plotted in Figure~\ref{fig:convergence}, which takes the validation perplexity as y-axis and the training wall time as x-axis. 
Although larger dense models achieve better validation perplexity at last, their training speed is quite slow. 
With regard to the convergence speed, MoE-based Transformers usually exceed dense models. 
Further, among the MoE methods, \ours{} has the fastest convergence speed. 

\subsubsection{Multilingual Machine Translation}

% [MT main results. + table]
We compare \ours{} with Switch Transformer, BASE Layer, Hash Layer, the standard Transformer, and a larger Transformer. 
All MoE-based models have the same number of shared parameters as the standard Transformer. 
Except the standard Transformer, the other models have the same FLOPs. 

We translate other languages to English~(X$\rightarrow$En) and report the test BLEU on WMT in Table~\ref{tab:main_nmt}. 
\ours{} achieves the best average test BLEU among the compared MoE methods. 
Keeping the same FLOPs, \ours{} outperforms the dense model by 1.22 test BLEU. 
With the MoE technique, we expand the number of parameters by 523\% and the FLOPs just increase by 9.3\%. 

\subsection{Analysis}

\subsubsection{Effects of Hyperparameters}

On top of the base setting of language modeling, we investigate different settings for the MoE layers in \ours{}. 

\begin{figure}[t]
\centering
\includegraphics[width=0.92\linewidth]{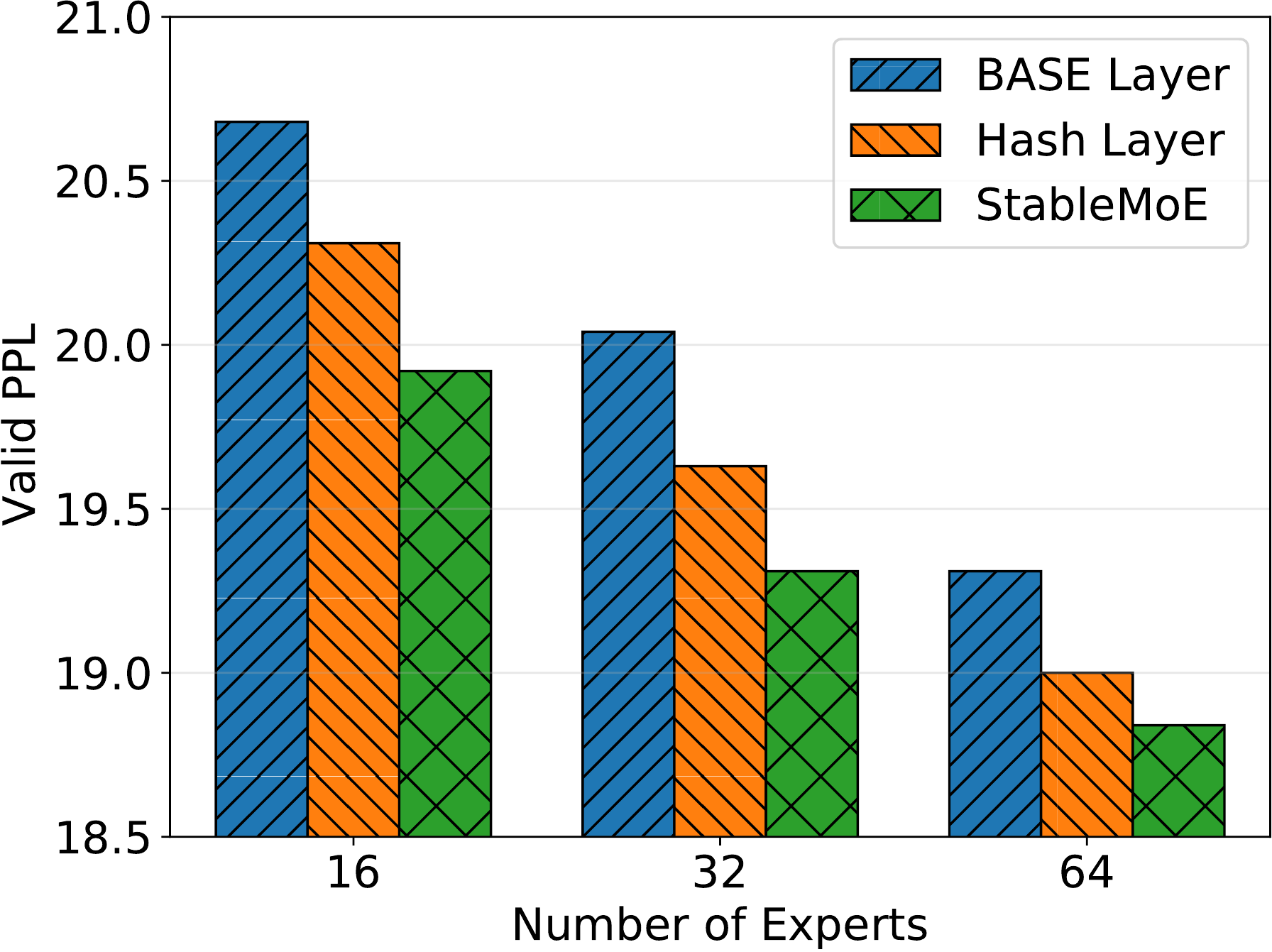}
\caption{
Comparison of MoE-based Transformers with different numbers of experts.
Lower perplexity indicates better performance.
}
\label{fig:expert_num}
\end{figure}

% [expert number. + figure]
\paragraph{Number of Experts}
Figure~\ref{fig:expert_num} shows the results of BASE Layer, Hash Layer, and \ours{} with different numbers of experts. 
As the number of experts goes larger, the validation perplexity of each model tends to further descend. 
Consistently, \ours{} performs the best with different numbers of experts. 
In addition, it is worth noting that \ours{} with 16 experts outperforms BASE Layer with 32 experts, and \ours{} with 32 experts achieves a similar perplexity to BASE Layer with 64 experts. 

\begin{figure}[t]
\centering
\includegraphics[width=0.92\linewidth]{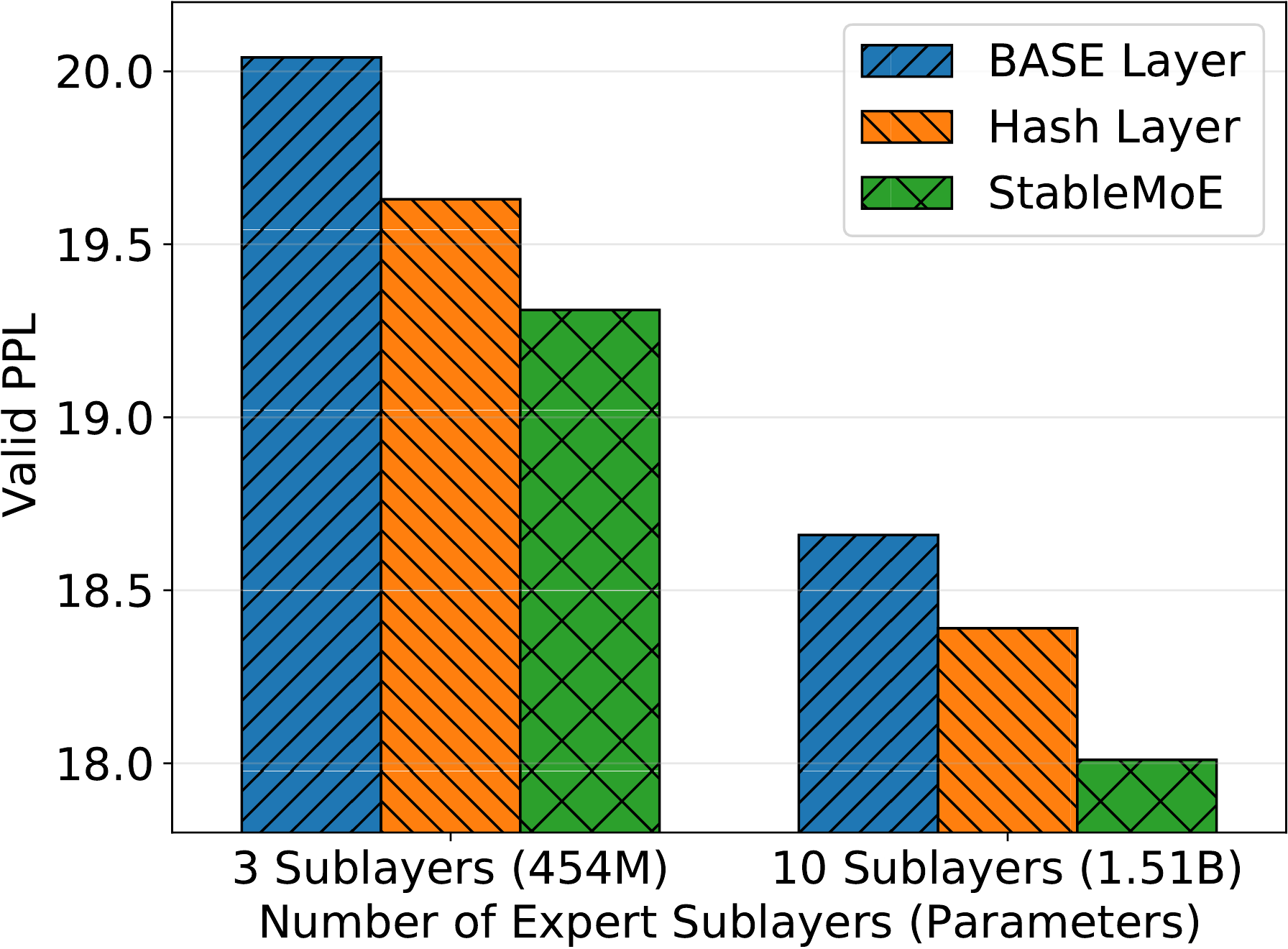}
\caption{
Comparison of MoE models with different numbers of expert sublayers (i.e., number of parameters).
Lower perplexity indicates better performance.
}
\label{fig:expert_param}
\end{figure}

% [MoE parameter number. + table]
\paragraph{Number of Expert Parameters}
We compare MoE models with different numbers of expert parameters by setting different expert sublayers. 
Models with 3 and 10 expert sublayers have 454M and 1.51B expert parameters, respectively. 
From Figure~\ref{fig:expert_param}, we observe that more expert parameters bring better performance, and \ours{} consistently performs the best under both settings. 

\begin{table}[t]
\centering
\footnotesize
\setlength{\tabcolsep}{25pt}
\begin{tabular}{@{}l c@{}}
\toprule
\textbf{Models} & \textbf{Valid PPL} \\
\midrule
\ours{} (stacked, top) & 19.55 \\
\ours{} (stacked, middle) & \textbf{19.28} \\
\ours{} (stacked, bottom) & 22.82 \\
\midrule
\ours{} (scattered) & 20.56 \\
\bottomrule
\end{tabular}
\caption{
Effects of the position of MoE layers. 
\ours{} (scattered) scatters 3 MoE sublayers uniformly into the standard Transformer, while the others stack 3 MoE sublayers together. 
}
\label{tab:expert_layer_num_pos}
\end{table}

% [MoE layer position. + table]
\paragraph{Position of MoE Layers}
We investigate the effect of the inserting position of the MoE layer. 
By default, the MoE layer stacks 3 MoE sublayers and is inserted after the $\frac{L}{2}$-th Transformer block (middle). 
We also attempt to insert the MoE layer before the first Transformer block (bottom), and after the last Transformer block (top). 
In addition, we also investigate the effect if we scatter 3 MoE sublayers uniformly into the standard Transformer, i.e., after the $\frac{L}{4}$-th, $\frac{2L}{4}$-th, and $\frac{3L}{4}$-th blocks, respectively. 
As shown in Table~\ref{tab:expert_layer_num_pos}, among the above four settings, inserting stacked MoE sublayers into the middle position allows \ours{} to achieve the best performance.

\paragraph{Ratio Between Two Training Stages}
We investigate the balance point of the ratio between two training stages in \ours{}. 
Given a fixed number of total steps, allocating more steps to training stage 1 can help to learn and distill a better routing strategy. 
On the other hand, a larger ratio of training stage 2 means longer stable training. 
Under the base setting of language modeling, we attempt to allocate 6K, 15K, and 30K steps to training stage 1 and show the results in Table~\ref{tab:distill}.
We find that if we use word embeddings as the distilled router, allocating 6K steps (10\% of the total steps) to training stage 1 is a good balance point. 
We speculate that the word embedding is simple enough to be learned fast, so longer stable training is more important to achieve better performance.

\begin{table}[t]
\centering
\footnotesize
\setlength{\tabcolsep}{10pt}
\begin{tabular}{@{}l l@{}}
\toprule
\textbf{Models} & \textbf{Valid PPL} \\
\midrule
BASE Layer & 20.04 \\
~~ $+$ Fixed Routing Strategy (Stage 2) & 19.41~(0.63$\downarrow$) \\
\midrule
\ours{} with Only Stage 1 & 19.48 \\
~~ $+$ Fixed Routing Strategy (Stage 2) & 19.28~(0.20$\downarrow$) \\
\bottomrule
\end{tabular}
\caption{
Effects of the fixed routing strategy. 
}
\label{tab:ablation}
\end{table}

\subsubsection{Effects of the Fixed Routing Strategy}

Based on the base setting of language modeling, we design two experiments to investigate how much performance improvement the fixed routing strategy can bring. 
On the one hand, we equip BASE Layer with a stable routing strategy to address its routing fluctuation problem. 
Specifically, as in \ours{}, we use word embeddings to distill the routing strategy of BASE Layer in the first 6K training steps, and freeze the distilled router for stable routing in the remaining training. 
As shown in Table~\ref{tab:ablation}, the fixed routing strategy decreases the validation perplexity of BASE Layer by 0.63. 
On the other hand, we attempt to disable the training stage 2 in \ours{} and always train the model as in training stage 1. 
As a result, the validation perplexity of \ours{} becomes 0.20 higher than the full version that has a fixed routing strategy. 
These two cases support that the fixed routing strategy, which addresses the routing fluctuation problem, can bring better performance for MoE-based Transformers. 

In addition, we visualize the fixed routing strategy of \ours{} in Appendix~\ref{appendix:visualization} for reference. 

\begin{table}[t]
\centering
\footnotesize
\setlength{\tabcolsep}{10pt}
\begin{tabular}{@{}l c c@{}}
\toprule
\textbf{Distilled Routers} & \textbf{Stage 1 Steps} & \textbf{Valid PPL} \\
\midrule
Word Embedding & 6K~(10\%) & \textbf{19.28} \\
Word Embedding & 15K~(25\%) & 19.34 \\
Word Embedding & 30K~(50\%) & 19.41 \\
\midrule
CNN & 15K~(25\%) & 19.39 \\
1-layer Transformer & 15K~(25\%) & 19.42 \\
2-layer Transformer & 15K~(25\%) & 19.38 \\
3-layer Transformer & 15K~(25\%) & 19.65 \\
\bottomrule
\end{tabular}
\caption{
Results of different ratios of two training stages and different variants of distilled routers. 
}
\label{tab:distill}
\end{table}

\subsubsection{Variants of Distilled Routers}
\label{sec:distill}

In Table~\ref{tab:distill}, in addition to word embedding, we also investigate four variants of the distilled router including CNN and three Transformers with different numbers of layers. 
We allocate 15K steps to training stage 1 for all of them. 
From the table, we find that using word embedding achieves the best performance, while the 3-layer Transformer does not perform well. 
For the routing strategy distillation, the distilling signal from a 32-category classification objective may not be informative enough to learn a complex router. 
By contrast, it is more suitable for simpler routers. 
Therefore, we recommend using word embedding, which is simple and effective, as the distilled router in \ours{}. 

\subsubsection{Analysis of Routing Fluctuations}

\begin{figure}[t]
\centering
\includegraphics[width=0.99\linewidth]{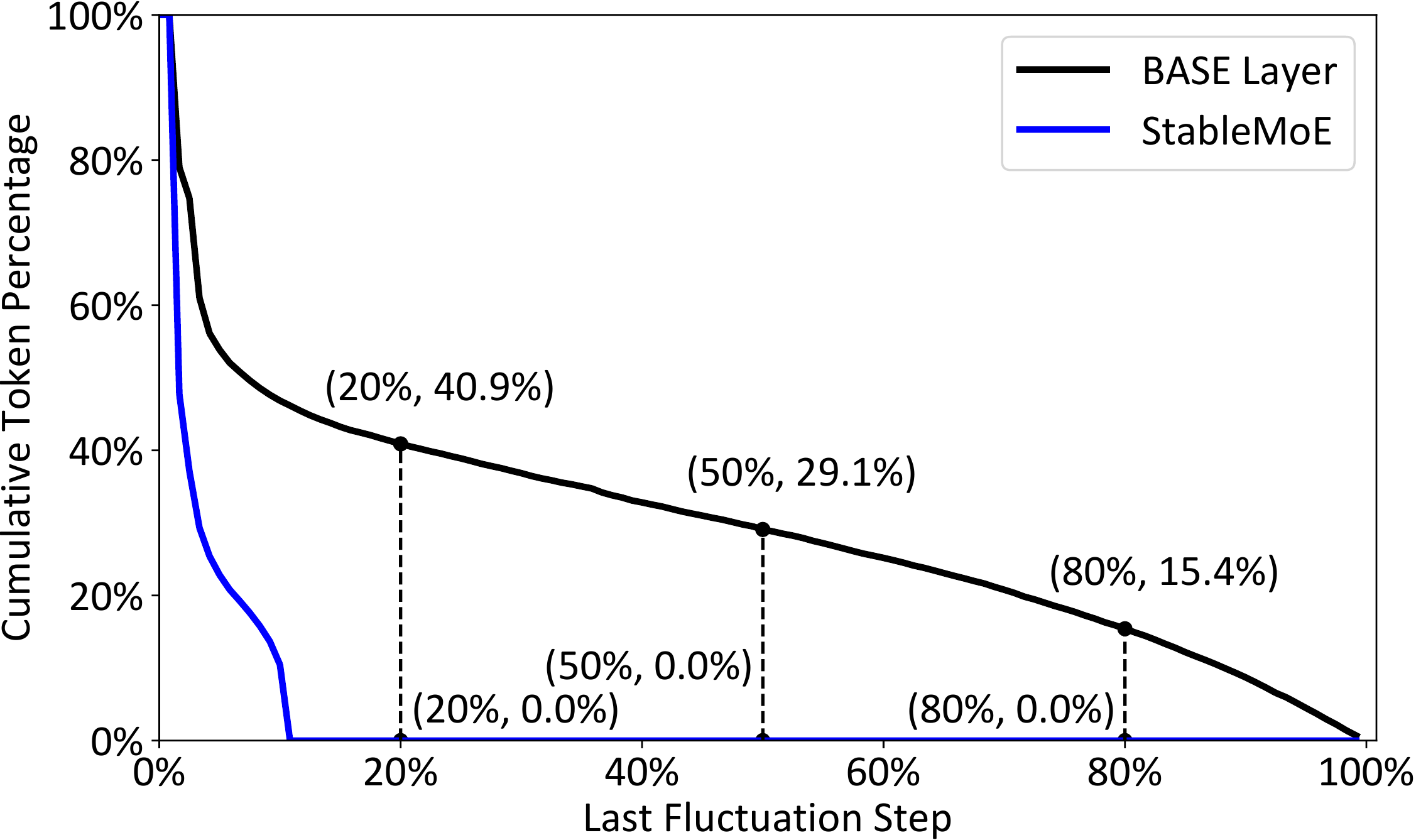}
\caption{
Cumulative token percentage about the last fluctuation step of tokens for BASE Layer and \ours{}.
Notice that training stage 2 of \ours{} does not have routing fluctuation compared with BASE Layer. 
}
\label{fig:last_fluctuation_cp}
\end{figure}

We compare the degree of routing fluctuations between \ours{} and BASE Layer to show our advantage with regard to the routing stability.  
During the 60K training steps, we examine the token-to-expert assignment for tokens in the validation set every 500 steps. 
For each token, we define the last fluctuation step as the last step where its target expert is different from the final step. 
We plot the cumulative token percentage about the last fluctuation step in Figure~\ref{fig:last_fluctuation_cp}. 
For ease of reading, we annotate the x-axis as the percentage it accounts for all training steps. 
From the figure, we find that the routing fluctuation problem is notable for BASE Layer. 
By contrast, for \ours{}, there is no routing fluctuation in training stage 2 since we apply a fixed routing strategy.

\section{Related Work}

\citet{ori_moe1,ori_moe2} propose Mixture of Experts~(MoE) to compute different examples with independent expert modules. 
\citet{moe} introduce MoE to build large-scale language models based on LSTMs~\citep{lstm}. 
Recently, as Transformers become popular, many pieces of work design MoE-version FFNs to build MoE-based Transformers. 
GShard~\citep{gshard}, Switch Transformer~\citep{switch}, and BASE Layer~\citep{base} follow the learning-to-route paradigm and dynamically learn how to route each input token to experts. 
However, we point out that these learning-to-route methods face the routing fluctuation problem. 
Hash Layer~\citep{hash} propose a non-parametric routing strategy, which uses a pre-designed token-level hash table to determine the token-to-expert assignment. 
The static routing strategy will not fluctuate, but the randomly determined hash table limits the upper bound of its performance. 
Our work includes the advantages of learning-to-route methods to learn a balanced and cohesive routing strategy, and further addresses the routing fluctuation problem through applying a frozen lightweight router that mimics the original routing strategy.

\section{Conclusion}

In this paper, we point out the routing fluctuation problem that exists in previous learning-to-route MoE methods. 
In order to address this problem, we propose \ours{} with two training stages. 
We first learn a balanced and cohesive routing strategy and synchronously distill it into a lightweight router decoupled from the backbone model. 
Then, we freeze the distilled router for a stable routing strategy in the remaining training. 
We validate \ours{} on language modeling and multilingual machine translation. 
The results show that \ours{} outperforms existing MoE methods in terms of both convergence speed and performance.

% \section*{Acknowledgement}

% Damai Dai, Zhifang Sui, and Baobao Chang are supported by the National Key Research and Development Program of China 2020AAA0106701 and NSFC project U19A2065. 

\bibliography{anthology,custom}
\bibliographystyle{acl_natbib}

\newpage
\appendix

\section*{Appendix}

\section{Hyper-parameters for Language Modeling}
\label{appendix:hyper_lm}

The hyper-parameters of \ours{} under the base and the large settings for language modeling are summarized in Table~\ref{tab:hyper_lm}. 

\begin{table}[h]
\centering
\footnotesize
\setlength{\tabcolsep}{2pt}
\begin{tabular}{@{}l c c@{}}
\toprule
\textbf{Hyper-parameters} & \textbf{Base} & \textbf{Large} \\
\midrule
Number of Experts & 32 & 64 \\
Number of MoE Layers & 1 & 1 \\
Sublayers per Expert & 3 & 6 \\
Embedding \& Hidden Size & 768 & 1024 \\
FFN Inner Hidden Size & 3072 & 4096 \\
Number of Attention Heads & 12 & 16 \\
Number of Transformer Blocks & 12 & 24 \\
\midrule
Sequence Length & 1024 & 1024 \\
Batch Size & 512K Tokens & 512K Tokens \\
\midrule
Optimizer & Adam & Adam \\
Maximum Learning Rate & 6e-4 & 3e-4 \\
Learning Rate Scheduler & Linear Decay & Linear Decay \\
Total Steps & 60K & 60K \\
Warm-up Steps & 2K & 2K \\
Gradient Clip Norm & 0.1 & 0.1 \\
Dropout & 0 & 0 \\
\bottomrule
\end{tabular}
\caption{
Hyper-parameters of \ours{} under the base and the large settings for language modeling. 
}
\label{tab:hyper_lm}
\end{table}

\section{Hyper-parameters for Multilingual Machine Translation}
\label{appendix:hyper_mt}

The hyper-parameters of \ours{} for multilingual machine translation are summarized in Table~\ref{tab:hyper_mt}. 

\begin{table}[h]
\centering
\footnotesize
\setlength{\tabcolsep}{8pt}
\begin{tabular}{@{}l c c@{}}
\toprule
Number of Experts & 32 \\
Number of MoE Layers & 2 \\
Sublayers per Expert & 3 \\
Embedding \& Hidden Size & 512 \\
FFN Inner Hidden Size & 2048 \\
Number of Attention Heads & 8 \\
Number of Transformer Encoder Blocks & 6 \\
Number of Transformer Decoder Blocks & 6 \\
\midrule
Maximum Sequence Length & 256 \\
Maximum Batch Size & 512K Tokens \\
\midrule
Optimizer & Adam \\
Maximum Learning Rate & 5e-4 \\
Learning Rate Scheduler & InvSqrt \\
Total Steps & 352K \\
Warm-up Steps & 4K \\
Gradient Clip Norm & 0.1 \\
Dropout & 0.1 \\
Attention Dropout & 0 \\
Label Smoothing & 0.1 \\
\bottomrule
\end{tabular}
\caption{
Hyper-parameters of \ours{} for multilingual machine translation. 
}
\label{tab:hyper_mt}
\end{table}

\begin{table*}[t]
\centering
\footnotesize
\setlength{\tabcolsep}{4pt}
\begin{tabular}{l l l}
\toprule
\textbf{Experts} & \textbf{Most Frequent Tokens} & \textbf{Descriptions} \\
\midrule
5 & my, his, her, year, years, day, life, week, family, days  & possessive case \& time units \\
6 & with, at, from, about, them, need, want, him, against, using & prepositions \& objective case \\
11 & that, ?, !, which, )., .", That, "., .), !!, ?", !!!, :), Â, !", ?, !, !), & conjunctions \& punctuations \\
12 & one, what, some, any, two, many, \$, use, 2, 1 & numerals \\
13 & information, support, experience, service, data, services, money, access, research & nouns about technologies \\
17 & world, government, state, country, community, city, 2018, United, US, law & nouns about politics \\
22 & right, business, high, free, important, public, big, top, hard, small & adjectives \\
27 & time, work, home, place, care, water, area, health, job, car & nouns about the daily life \\
29 & ing, a, ed, in, er, on, o, e, as, es, an, al, en, am, it, is, ie, os, le & suffixes \\
30 & you, we, they, there, It, We, here, You, ve, 've & pronouns \\
31 & and, or, by, when, after, through, before, while, And, until & conjunctions \\
\bottomrule
\end{tabular}
\caption{
The most frequent tokens assigned to each expert in the validation set. 
We present several representative experts. 
Tokens assigned to the same expert usually share some common features. 
}
\label{tab:cohesive}
\end{table*}

\section{Visualization of the Fixed Routing Strategy of \ours{}}
\label{appendix:visualization}

We visualize the fixed routing strategy of \ours{} in Table~\ref{tab:cohesive}. 
On the validation set, for each expert, we demonstrate the most frequent tokens assigned to it along with a text that describes their common features.  
We find that tokens assigned to the same expert usually share some common features, e.g., Expert 22 captures adjectives and Expert 31 captures conjunctions.
These cases show good cohesiveness of the token-to-expert assignment in \ours{}. 

\end{document}

%% file: settings.tex
\usepackage{multirow}
\usepackage{amsmath}
\usepackage{capt-of}
\usepackage{tabularx}
\usepackage{epsfig}
\usepackage{amssymb}
\usepackage{amsfonts}
\usepackage{booktabs}
\usepackage{scalerel}
\usepackage[inline]{enumitem}
\usepackage{listings}
\usepackage{varwidth}
\usepackage[export]{adjustbox}
\usepackage{tikz}
\usetikzlibrary{tikzmark}
\usepackage{cleveref}

\usepackage{stmaryrd}
\usepackage{bbm}

\usepackage{algorithm}
\usepackage[noend]{algpseudocode}

\definecolor{deepblue}{rgb}{0,0,0.5}
\definecolor{officeblue}{RGB}{0,102,204}
\definecolor{deepred}{rgb}{0.6,0,0}
\definecolor{deepgreen}{rgb}{0,0.5,0}
\definecolor{mybrickred}{RGB}{182,50,28}

\definecolor{fillcolor}{RGB}{216,217,252}

\algnewcommand\algorithmicrequireb{{\hspace{0.85cm}}}
\algnewcommand\INPTDESCB{\item[\algorithmicrequireb]}

\algnewcommand\algorithmicfuncdesc{\textbf{Function:}}
\algnewcommand\FUNCDESC{\item[\algorithmicfuncdesc]}
\algnewcommand\algorithmicfuncdescb{{\hspace{1.48cm}}}
\algnewcommand\FUNCDESCB{\item[\algorithmicfuncdescb]}
\algnewcommand{\algorithmicgoto}{\textbf{goto}}
\algnewcommand{\Goto}[1]{\algorithmicgoto~\ref{#1}}

%% file: math_commands.tex
\usepackage{amsmath,amsfonts,bm}

\def\eqref#1{equation~\ref{#1}}
\def\1{\bm{1}}

\DeclareMathAlphabet{\mathsfit}{\encodingdefault}{\sfdefault}{m}{sl}
\SetMathAlphabet{\mathsfit}{bold}{\encodingdefault}{\sfdefault}{bx}{n}

\newcommand{\sigmoid}{\sigma}